\documentclass[letterpaper]{article} 
\usepackage{aaai2027}  
\usepackage[hyphens]{url}  
\usepackage{graphicx} 
\usepackage{natbib}  
\usepackage{caption} 
\usepackage{algorithm}
\usepackage{algorithmic}

\usepackage{newfloat}
\usepackage{amsfonts} 
\usepackage{listings}
\DeclareCaptionStyle{ruled}{labelfont=normalfont,labelsep=colon,strut=off} 
\floatstyle{ruled}
\newfloat{listing}{tb}{lst}{}
\floatname{listing}{Listing}

\usepackage{booktabs}

\usepackage{graphicx}
\usepackage{booktabs}
\usepackage[table]{xcolor}
\usepackage{multirow}
\usepackage{enumitem}
\usepackage{amsmath}

\definecolor{mypurple}{RGB}{126, 100, 158}
\definecolor{mygreen}{RGB}{120, 148, 64}
\definecolor{morandiblue}{HTML}{DCE6F2}
\definecolor{berry}{RGB}{204, 0, 102}

\title{MMPhysVideo: Physically Plausible Video Generation Through Joint RGB–Perception Modeling}
\author{
    Shubo Lin\textsuperscript{\rm 1,2,3,4}, Xuanyang Zhang\textsuperscript{\rm 3}\thanks{Project Lead.}, Wei Cheng\textsuperscript{\rm 3}, Weiming Hu\textsuperscript{\rm 1,2,4,5}, Gang Yu\textsuperscript{\rm 3}\corresponding, Jin Gao\textsuperscript{\rm 1,2,4}\corresponding\\
}
\affiliations{
    \textsuperscript{\rm 1}State Key Laboratory of Multimodal Artificial Intelligence Systems, CASIA\\
    \textsuperscript{\rm 2}School of Artificial Intelligence, University of Chinese Academy of Sciences\\
    \textsuperscript{\rm 3}StepFun, \textsuperscript{\rm 4}Beijing Key Laboratory of Super Intelligent Security of Multi-Modal Information\\
    \textsuperscript{\rm 5}School of Information Science and Technology, Shanghai Tech University

}

\begin{document}

\maketitle

\begin{abstract}
Despite advancements in generating visually stunning content, video diffusion models (VDMs) often yield physically inconsistent results due to pixel-only reconstruction.
  To address this, we propose MMPhysVideo, the first study to enhance physical plausibility in video generation through joint multimodal modeling. We recast perceptual cues, specifically semantics, geometry, and spatio-temporal trajectories, into a unified pseudo-RGB format, enabling VDMs to directly capture complex physical dynamics.
  To mitigate cross-modal interference, we propose a Bidirectionally Controlled Teacher architecture, which utilizes parallel branches to fully decouple RGB and perception processing and adopts two zero-initialized control links to gradually establish pixel-wise consistency. For inference efficiency, the teacher’s physical prior is distilled into a single-stream student model via representation alignment. Furthermore, we present MMPhysPipe, an end-to-end data curation and annotation pipeline tailored for constructing physics-rich multimodal datasets.
  MMPhysPipe employs a vision-language model (VLM) guided by a chain-of-visual-evidence rule to pinpoint physical subjects, enabling expert models to extract multi-granular perceptual information.
  Without additional inference costs, MMPhysVideo consistently improves physical plausibility of advanced models on the Videophy and PhyGenbench benchmarks and achieves superior performance among existing methods. The project page of MMPhysVideo is at \url{https://shubolin028.github.io/MMPhysVideo-Page}.
\end{abstract}


\section{Introduction}
\label{sec:intro}

Fueled by the scaling of compute and data, video generation models have advanced rapidly~\cite{hunyuanvideo1p5,opensora2p0}, enabling a wide range of applications in content creation~\cite{vace,dreamvideo} and embodied agents~\cite{robomaster,taste_rob}. Despite remarkable progress in visual aesthetics and fidelity, the lack of physical plausibility in synthesized videos remains a critical bottleneck in the pursuit of real-world simulators.

To align visual outputs with physical principles, current studies focus on the integration of physical priors into text-to-video (T2V) generators. One technical route encodes structured physical priors as prompts~\cite{phyt2v,wisa} or guidance~\cite{wmreward,phyrpr} for T2V generators. However, a significant domain gap exists when mapping textual physical descriptions to pixel-level distribution sampling~\cite{prophy}. Another technical route adopts a post-tuning paradigm, which implicitly transfers knowledge from foundation models into generators via supervised fine-tuning (SFT)~\cite{prophy,videorepa} or reinforcement learning (RL)~\cite{physhpo,phygdpo,physrvg}. For instance, VideoREPA~\cite{videorepa} aligns the representations of self-supervised foundation models like DINO~\cite{dinov2} or VideoMAE~\cite{videomae,videomaev2} with video diffusion models (VDMs) through distillation-based SFT. While effective, the relation between self-supervised representations and abstract physics principles remains tenuous. Alternatively, PhysHPO~\cite{physhpo} introduces hierarchical preference optimization guided by vision–language models (VLMs)~\cite{qwen2}. Nonetheless, such RL-based methods rely on complex reward design, which often suffer from reward hacking~\cite{videoalign} and prohibitive training costs~\cite{phygdpo,unireward}. These limitations raise a question: \textit{Is there a more robust mechanism for integrating physical priors into video generators that ensures explicit correspondence with specific physical content?}

\begin{figure*}[t!]
    \centering
    \includegraphics[width=0.98\textwidth]{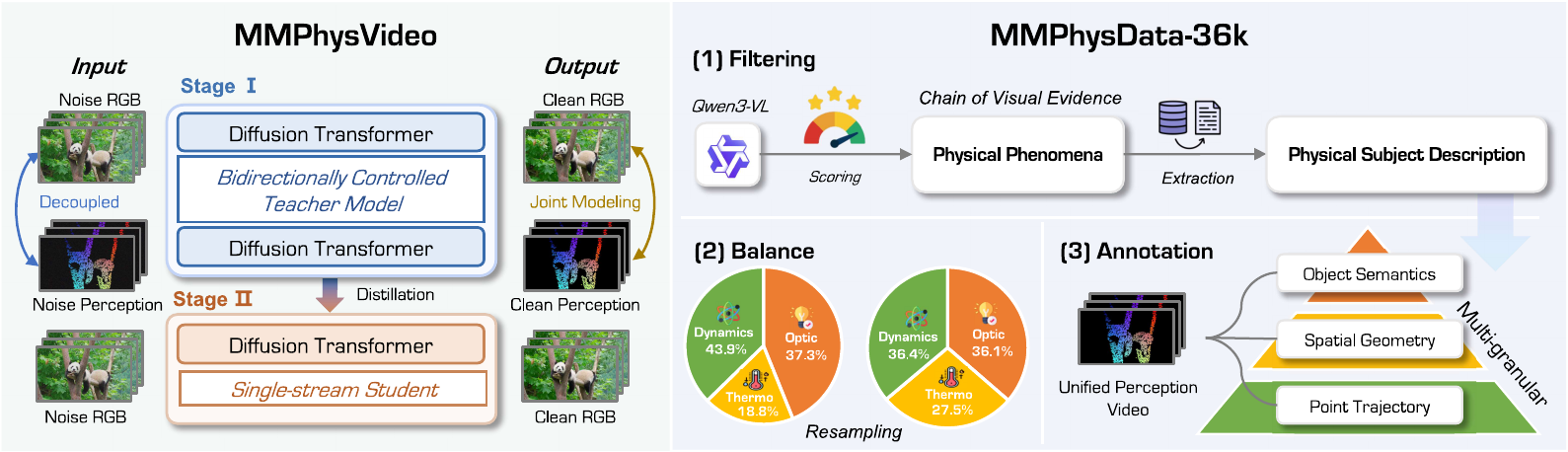}
    \vspace{-2pt}
    \caption{\textbf{Overall framework of MMPhysVideo.} \textit{Left}: Our two-stage training framework, which starts with teacher models of parallel branches for joint modeling. Then, we distill a single-stream student model through representation alignment. \textit{Right}: Our data engine, MMPhysPipe, for physics data curation and multimodal annotation.}
    \label{fig:workflow_overall}
    \vspace{-10pt}
\end{figure*}

We answer this question affirmatively by recasting perceptual cues, e.g. segmentation, geometry, as pseudo-RGB frames. This strategy provides a data format inherently compatible with VDMs,  empowering them to perform generative world modeling after mastering the perception and understanding of the underlying physical world.
To achieve this goal, we identify and target two challenges: 

\begin{itemize}[topsep=0pt, partopsep=0pt, itemsep=0pt]
    \item Designing an effective \textcolor{mygreen}{\textbf{joint modeling architecture}} to enhance physical consistency in generated videos. While existing diffusion-based perception methods typically employ a single-stream architecture via channel-wise~\cite{geovideo} or spatial-wise~\cite{4dnex} concatenation, directly adopting such designs leads to suboptimal performance on physics-centric benchmarks (Tab.~\ref{tab:aba_arch_percep}).
    \item Exploiting \textcolor{mypurple}{\textbf{physics-rich multimodal datasets}}. Unlike classic post-tuning methods~\cite{physhpo} or conditional generation tasks~\cite{magicmotion}, our joint world modeling requires real-world videos with both rich physical phenomena and paired perceptual annotations capturing multi-granular physical dynamics.
\end{itemize} 

\noindent To tackle these challenges, we propose \textbf{MMPhysVideo}, a framework designed to enhance the physical plausibility of video generation through joint multimodal training. MMPhysVideo starts with a Bidirectionally Controlled Teacher (BCT) model that processes RGB and perceptual modalities in a decoupled manner. To minimize inter-modal interference, we assign the two modalities to weight-shared parallel branches, ensuring that each modality utilizes an independent computational stream. We then incorporate bidirectional control links with zero-initialization to gradually learn pixel-wise alignment between RGB and perceptual modalities. To avoid additional computational overhead, the teacher model is distilled into a single-stream student via relation-based representation alignment.

Furthermore, we propose a physics-centric data curation and multimodal annotation pipeline, \textbf{MMPhysPipe}. Concretely, MMPhysPipe employs a VLM to categorize the physical phenomena in each video and generate the corresponding subject descriptions, which provide a chain-of-visual-evidence reasoning rule for the VLM to assess the physical richness score of each video. Subsequently, MMPhysPipe leverages the grounding capabilities of \textit{Segment Anything Model} (SAM)~\cite{sam3} to translate these subject descriptions into semantic masks. Finally, to capture spatial geometry and fine-grained dynamics, MMPhysPipe performs pointmap (XYZ) annotation~\cite{vggt} and 3D point tracking~\cite{spatialtrackerv2} within the masked regions. This multimodal mix effectively provides the dense perception supervision necessary for a world model to generate consistent physical interactions.

In summary, our main contributions are as follows: 
\textbf{(1)} We introduce MMPhysVideo, the first post-tuning framework that enhances the physical plausibility of video synthesis by joint multimodal modeling. MMPhysVideo adopts a Bidirectionally Controlled Teacher (BCT) architecture to preserve video priors while effectively learning physical perception dynamics. To maintain inference efficiency, we distill the teacher's prior into a single-stream student via relation-based representation alignment.
\textbf{(2)} We present MMPhysPipe, a comprehensive data curation and annotation pipeline to construct MMPhysData-36k with physics-rich multimodal datasets.
\textbf{(3)} Quantitative and qualitative results show that MMPhysVideo brings remarkable improvements in physical plausibility over various advanced models and surpasses state-of-the-art methods on physics-centric benchmarks. 

\section{Related Works}

\noindent \textbf{Physics-aware video generation.} Benefiting from scaled data~\cite{opensora2p0} and advanced architectures~\cite{dit}, video diffusion models (VDMs)~\cite{worldsimbench} have made significant strides in visual fidelity and sensory realism~\cite{hunyuanvideo1p5,lavie}. However, physical consistency remains a formidable challenge even for state-of-the-art models~\cite{2501generative,worldmodelbench}. While existing methods~\cite{physgaussian,physctrl} leverage physics simulators to generate physics-grounded videos, this technical route is often bottlenecked by the simulator’s narrow scope and the inherent sim-to-real gap. Recent research has shifted toward integrating physical priors within VDMs. One paradigm treats physical knowledge as prompts~\cite{phyt2v} or guidance~\cite{newtongen,wmreward} to activate the physics potential in VDMs. Another paradigm~\cite{phygdpo,physhpo,prophy} utilizes post-tuning to transfer physical insights from foundation~\cite{dinov2} or specialized~\cite{qwen2} models to VDMs. This paper, from a distinct perspective, enables generative models to inherently internalize physical dynamics through joint RGB-perception modeling.

\noindent \textbf{Diffusion models for perceptual predictions.} Beyond synthesis, another line of works~\cite{percepdiff,diception} utilize diffusion models for various perception-related tasks. Notably, Marigold~\cite{marigold} pioneered the encoding of depth data as pseudo-RGB images to enable generative perceptual prediction. Following this idea, VDMs have been generalized by subsequent works to a broader range of perceptual modalities, including segmentation~\cite{vidseg}, surface normals~\cite{normalcrafter,diffrender}, and geometric estimation~\cite{geovideo,worldreel,one4d}. Recently, several works~\cite{videojam,unidiff,ommivdiff} have focused on joint mappings between perceptual conditions and video, achieving multimodal generation. However, naïve modality stacking often introduces inter-modal interference and degrades RGB quality. Different from these works, we investigate joint multimodal modeling as a means to improve physical consistency while preserving visual fidelity.

\section{Method}


\subsection{Preliminaries: Video Diffusion Models}

Video diffusion models synthesize content through an iterative denoising process guided by a learned data distribution. Our proposed MMPhysVideo is built upon latent diffusion models (LDMs)~\cite{ldm} with Diffusion Transformer (DiT)~\cite{dit}, which provide great model capacity for multimodal modeling. Given an input video $\mathbf{X}_{rgb} \in \mathbb{R}^{3 \times F \times H \times W}$, where $F$, $H$, $W$ denote the number of frames, height, and width, respectively, the 3D Variational Autoencoder (VAE) first projects it into a latent space $\mathbf{z}^{0}_{rgb} = \mathcal{E}(\mathbf{X}_{rgb})$, where $\mathbf{z}^{0}_{rgb} \in \mathbb{R}^{c \times f \times h \times w}$. Here, $c$ represents the embedding dimensions of 3D VAE, while $f$, $h$, and $w$ denote the compressed temporal duration, height, and width. During training, a timestep $t \sim \mathcal{U}(0,1)$ and Gaussian noise $\epsilon_{rgb} \sim \mathcal{N}(\mathbf{0},\mathbf{I})$ are sampled to produce a noisy latent $\mathbf{z}^{t}_{rgb}$ via $\mathbf{z}^{t}_{rgb}=\mathbf{z}^{0}_{rgb} + \sigma_t^2\epsilon_{rgb}$ with variance $\sigma_t^2$. 
The denoiser $u(\cdot;\theta)$ is optimized to minimize the noise estimation loss:
\begin{equation}
\mathcal{L} = \mathbb{E}_{\mathbf{z}^0_{rgb}, \epsilon, t} \left[ \| u(\mathbf{z}^t_{rgb}, y,t;\theta) - \epsilon_{rgb} \|^2 \right],
\end{equation}
where $y$ is the conditioning signal (e.g., text prompt), $\theta$ denotes the weights, and $u$ is the denoiser's prediction. During inference, the denoised latent $\hat{\mathbf{z}}^{0}_{rgb}$ is mapped back to pixel space via the VAE decoder $\mathcal{D}$ to yield the final video $\hat{\mathbf{X}}_{rgb}$.


\subsection{MMPhysVideo}
\label{sec:mmphysvideo}

MMPhysVideo extends video diffusion models (VDMs)~\cite{worldsimbench} toward joint RGB and perception modeling to better capture and understand the physical dynamics of the real world. Existing VDMs often suffer from artifacts such as semantic ambiguities, flickering scene structures, and inconsistent temporal deformations, all of which compromise the physical realism of generated videos. To address these issues, we incorporate three critical perceptual cues: semantics, geometry, and spatio-temporal trajectory.
To improve training efficiency, we consolidate these multimodal signals into a unified pseudo-RGB video representation, $\mathbf{X}_{percep} \in \mathbb{R}^{3 \times F \times H \times W}$, as detailed in Sec.~\ref{sec:mmphyspipe}. 

\subsubsection{Joint Multimodal Modeling}
Given a video sequence and its paired perceptual video $\mathbf{X}_{percep}$, we follow the LDMs paradigm to unify multimodal tasks:
\begin{equation}
\hat{\epsilon}_{rgb}, \hat{\epsilon}_{percep} = u(\mathbf{z}^t_{rgb},\mathbf{z}^t_{percep},y,t; \theta),
\end{equation}
where $\mathbf{z}^t_{percep}$ is the perceptual latent encoded by 3D VAE, $\hat{\epsilon}_{rgb}$ and $\hat{\epsilon}_{percep}$ represent the predicted noise for RGB and perceptual modalities, respectively. Unlike methods using perceptual cues as conditional inputs~\cite{motionprompt} or post-hoc estimates~\cite{depthcrafter}, our joint modeling paradigm generates both from noise to foster inherent perceptual understanding.
Our experiments in Sec.~\ref{sec:ablation} reveal that the model architecture for joint generation is crucial for overall performance. Consequently, we compare existing fusion strategies with our proposed decoupled architecture.

\begin{figure}[t!]
    \centering
    \includegraphics[width=0.98\linewidth]{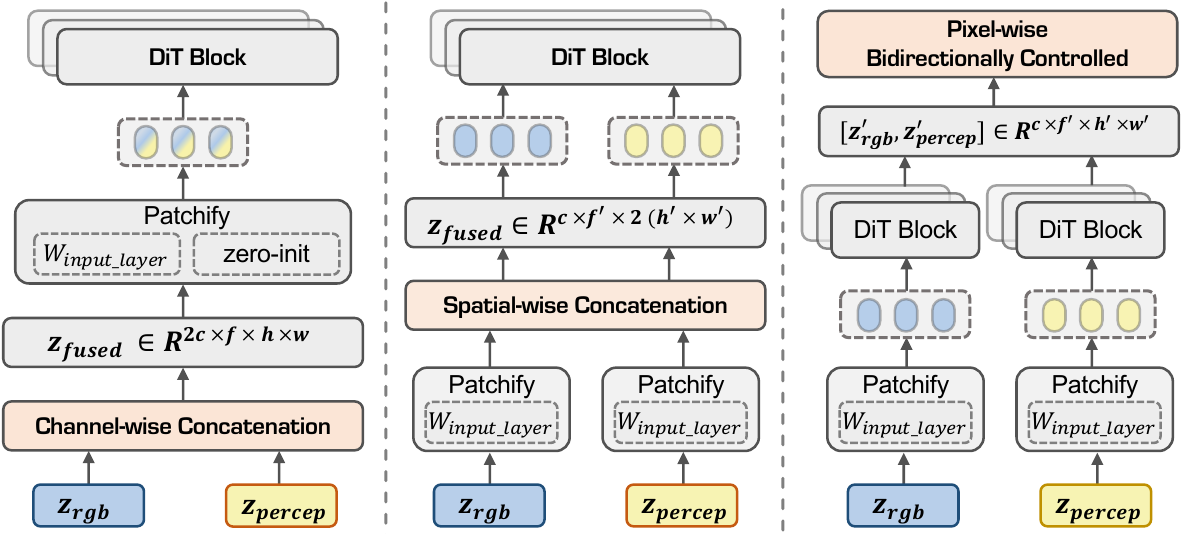}
    \vspace{-2pt}
    \caption{\textbf{Architecture comparison.} \textit{Left}: Channel-wise fusion~\cite{ommivdiff} \textit{Middle}: Spatial-wise fusion~\cite{unityvideo} \textit{Right}: Our pixel-wise fusion.}
    \label{fig:arch_comparison}
    \vspace{-10pt}
\end{figure}

\textit{Channel-wise concatenation.} Most attempts~\cite{videojam,unidiff} concatenate RGB and perceptual latents along the channel dimension, expanding input channels with zero-initialization. Although preserving sequence length, this early fusion may weaken perceptual representations and produce blurred outputs as shown in Fig.~\ref{fig:arch_vis_case}.

\textit{Spatial-wise concatenation.} Recent works~\cite{fulldit,ominicontrol} concatenate latents along spatial dimensions, independently patchifying each modality and encoding positions with 3D RoPE~\cite{roformer}. However, self-attention over non-aligned tokens induces redundant interactions, while the increased sequence length introduces additional computational overhead.

These \textbf{single-stream architectures} entangle heterogeneous modalities within a shared representation space, leading to inter-modal interference with suboptimal performance.

\begin{figure*}[t!]
    \centering
    \includegraphics[width=0.98\textwidth]{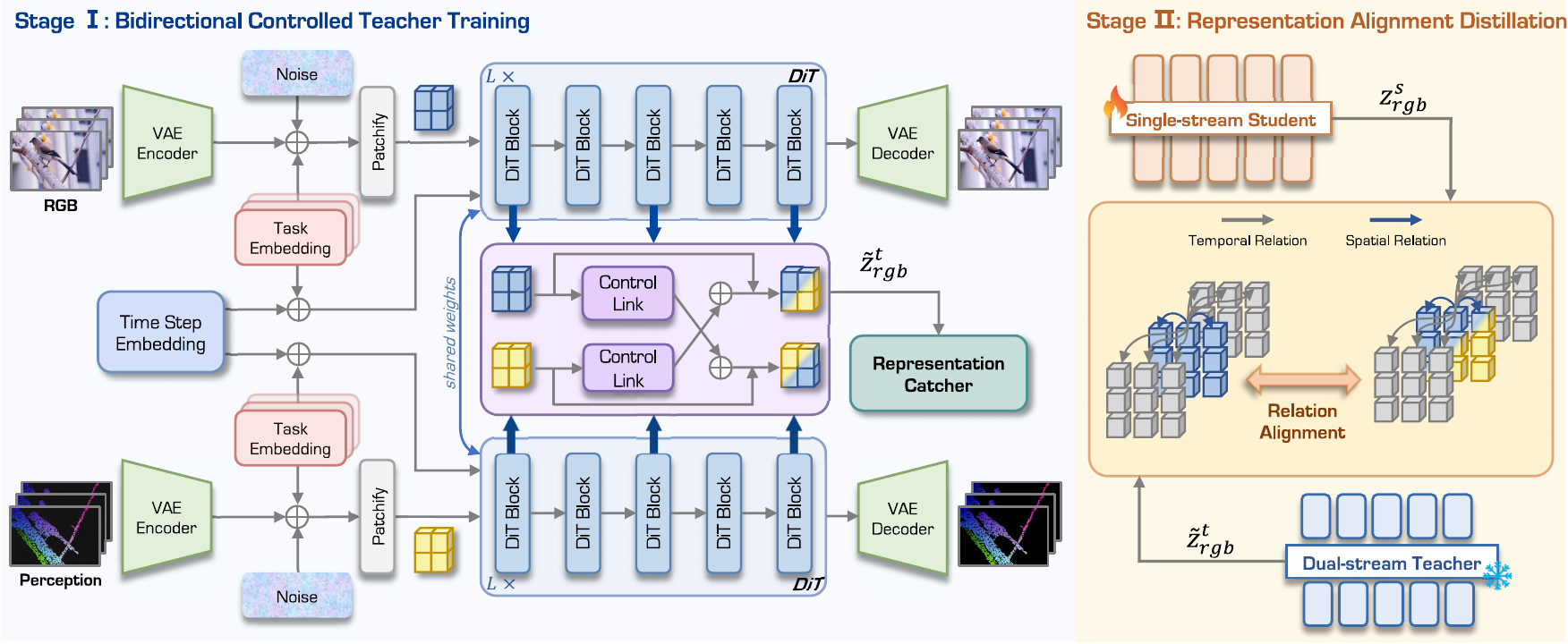}
    \caption{\textbf{Overview of MMPhysVideo.} \textit{Stage $I$}: A dual-stream teacher model with parallel branches is first trained to handle RGB and perceptual modalities concurrently. Then, we use bidirectional control links to enable pixel-wise alignment. \textit{Stage $II$}: For inference efficiency, we distill a single-stream student model through relation alignment.}
    \label{fig:workflow_mmphysvideo}
    \vspace{-10pt}
\end{figure*}

\subsubsection{Bidirectionally Controlled Teacher Model}
We address these issues in a divide-and-conquer manner. We first fully decouple the computation of RGB and perceptual modalities through parallel branches to prevent modal confusion. Subsequently, we introduce bidirectional control links to establish precise, pixel-wise correspondences between the modalities. 

\textit{Parallel branches.} Building upon standard RGB processing, we construct a parallel computational branch for the perceptual modality, forming a \textbf{dual-stream architecture}. Instead of duplicating all DiT parameters, two branches share model weights while keeping the forward computation independent. This design preserves the pre-trained RGB video priors while seamlessly extending VDMs to perceptual video generation. To further enforce modal discriminability, we introduce learnable task embeddings, $e_{rgb}$ and $e_{percep}$, which are integrated into the diffusion timestep embedding $e(t;\theta)$:
\begin{equation}
t_{rgb} = e_{rgb} + e(t;\theta), \quad t_{percep} = e_{percep} + e(t;\theta).
\end{equation}
Similarly, we also apply a distinct set of learnable task embeddings to the input latents. Then, the parallel computation branches can be formulated as:
\begin{equation}
\hat{\mathbf{z}}_{rgb} = {\rm DiTBlocks}(\mathbf{z}_{rgb}, t_{rgb}),
\end{equation}
\begin{equation}
\hat{\mathbf{z}}_{percep} = {\rm DiTBlocks}(\mathbf{z}_{percep}, t_{percep}).
\end{equation}
By isolating the computation, the DiT model can independently master the generation of the newly introduced perceptual modality, thereby mitigating optimization instability and reducing the overall learning difficulty.

\textit{Bidirectional control links.} To capture fine-grained physical dynamics, perceptual modalities such as 3D point tracking should maintain pixel-wise consistency with the RGB stream. Beyond merely training on paired data, we explicitly establish cross-modal alignment by introducing bidirectional control links within corresponding DiT blocks, which facilitates direct pixel-wise interaction between the RGB and perceptual representations.

For a DiT model consisting of $K$ blocks, we define a subset of indices $\mathcal{S} \subseteq \{1, \dots, K\}$ for link insertion. For $l$-th selected block ($l \in \mathcal{S}$), we inject the intermediate representations from one branch into the other to update the hidden states, $\tilde{\mathbf{z}}^{[l]}_{rgb}$ and $\tilde{\mathbf{z}}^{[l]}_{percep}$, which is formulated as:
\begin{equation}
\tilde{\mathbf{z}}^{[l]}_{rgb} = \mathbf{z}^{[l]}_{rgb} + {\rm ControlLink}_{percep \rightarrow rgb}(\mathbf{z}^{[l]}_{percep}),
\end{equation}
\begin{equation}
\tilde{\mathbf{z}}^{[l]}_{percep} = \mathbf{z}^{[l]}_{percep} + {\rm ControlLink}_{rgb \rightarrow percep}(\mathbf{z}^{[l]}_{rgb}),
\end{equation}
where each control link is implemented as a single linear layer by default. Following the convention of controllable generation~\cite{ctrlt2i}, we employ zero initialization for all control links. This strategy minimizes inter-modal interference during the early training phase and enables the model to progressively learn complex pixel-level correspondences between RGB and perceptual modalities.

\subsubsection{Representation alignment}
For inference efficiency, we distill the dual-stream teacher model back into a single-stream student model that processes only the RGB modality. However, directly performing MSE regression on raw features can be overly restrictive and degrade generation quality~\cite{reladistill}. Inspired by VideoREPA~\cite{videorepa}, we align the pairwise token similarities between the teacher’s and student’s representations, $\tilde{\mathbf{z}}^{t}_{rgb}$ and $\mathbf{z}^{s}_{rgb}$, across both spatial and temporal dimensions:
\begin{equation}
\begin{aligned}
\mathcal{L}_{distill} =
& \left| \mathcal{R}_{spa}(\tilde{\mathbf{z}}^{t}_{rgb})
-\mathcal{R}_{spa}(h_{\varphi}(\mathbf{z}^{s}_{rgb})) \right| \\
&+\left| \mathcal{R}_{temp}(\tilde{\mathbf{z}}^{t}_{rgb})
-\mathcal{R}_{temp}(h_{\varphi}(\mathbf{z}^{s}_{rgb})) \right|,
\end{aligned}
\label{eq:distill}
\end{equation}
where $\mathcal{R}_{spa}$ and $\mathcal{R}_{temp}$ denote functions that compute token relations in the spatial and temporal dimensions, respectively, and $h_{\varphi}$ is a trainable MLP projector. Through this objective, the student model learns the spatio-temporal structural dependencies inherent in the teacher’s fused RGB-perceptual representations. Initialized with teacher weights, the student model integrates dual-stream priors of perception and RGB into a single DiT. This allows the single-stream model to internalize physical priors and generate consistent videos without the auxiliary perceptual branch.

\begin{figure*}[t!]
    \centering
    \includegraphics[width=0.98\textwidth]{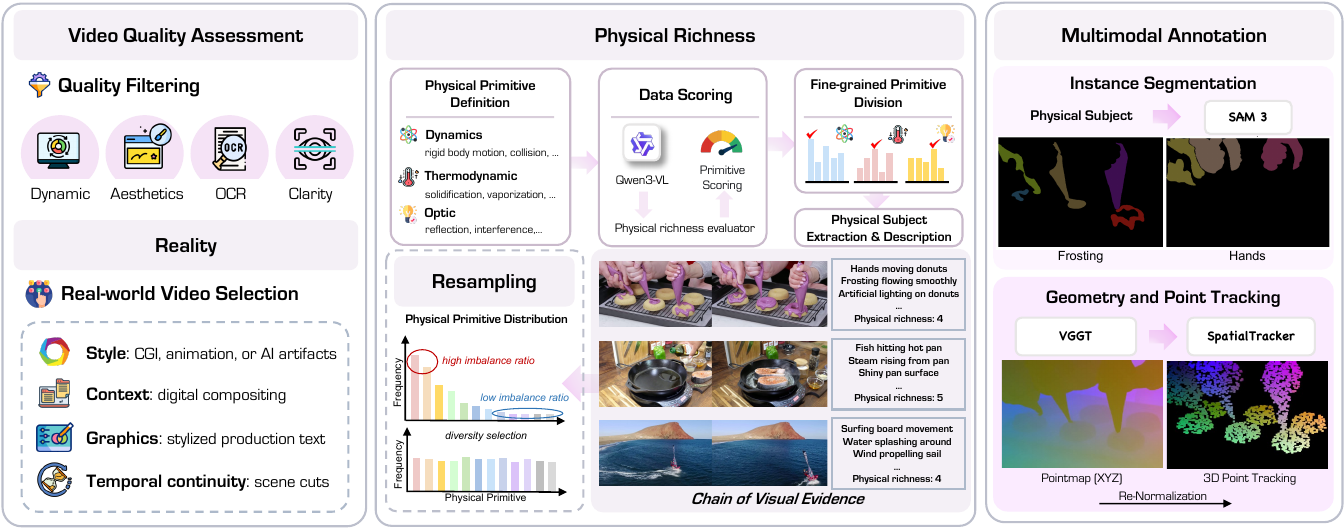}
    \caption{\textbf{Overview of MMPhysPipe.} We employ a VLM, Qwen3-VL~\cite{qwen3vl}, to curate videos with rich physical interactions and generate physical subject descriptions following our chain-of-visual-evidence (CoVE) rule. Subsequently, expert perception models~\cite{sam3,vggt,spatialtrackerv2} are leveraged to produce multi-granular annotations.}
    \label{fig:workflow_mmphyspipe}
    \vspace{-10pt}
\end{figure*}

\subsection{MMPhysPipe}
\label{sec:mmphyspipe}

To bridge the gap in physics-aware generation via joint multimodal modeling, we introduce MMPhysPipe, a scalable dataset pipeline designed to overcome two primary bottlenecks: the curation of video samples exhibiting rich physical phenomena from large text-video datasets and the production of high-fidelity, multi-granular perceptual annotations.

\subsubsection{Data Curation}

Given a large-scale text-video dataset~\cite{openvid}, MMPhysPipe first employs a lightweight Video Quality Assessment (VQA) model~\cite{koala} to efficiently filter out low-quality samples, such as freeze-frame or overexposed videos. The VQA model provides an initial holistic score for each video by integrating multiple dimensions, including dynamics, aesthetics, and clarity. 

\textit{Reality.} We further identify real-world captured videos using a Vision-Language Model (VLM). Specifically, we define real-world raw videos through four visual criteria: Style, Context, Graphics, and Temporal Continuity, which provide explicit constraints for Chain-of-Thought (CoT) reasoning.

\textit{Physical Richness.} Following prior works~\cite{wisa}, we categorize physical phenomena into three primary domains, each subdivided into fine-grained primitives. MMPhysPipe leverages Qwen3-VL~\cite{qwen3vl} as a physical evaluator to assign an intensity score ranging from $1$ to $5$ for each primitive, where $1$ denotes a negligible occurrence and $5$ represents high-impact manifestation. To mitigate hallucinated outputs, we implement a Chain-of-Visual-Evidence (CoVE)~\cite{vchain} reasoning rule: the VLM should provide visual evidence, such as momentum exchange in a collision, before scoring. 

Crucially, for primitives achieving the highest intensity scores, the VLM performs \textit{Physical Subject Extraction}, generating concise descriptive phrases (e.g., "steam rising from pan"). These phrases act as semantic anchors, bridging the gap between high-level physical principles and the subsequent low-level perceptual grounding annotation.

\textit{Resampling.}
To improve physical diversity, we introduce a multi-label balanced resampling strategy. 
Given the physical richness scores of all videos, we first convert primitive scores into binary labels according to a predefined threshold $\tau$. 
For primitives with imbalanced distributions, we assign larger sampling weights to under-represented primitives based on their inverse frequencies. 
Each video is then assigned a sampling probability according to the aggregated weights of its associated physical primitives. 
Finally, we sample videos according to these probabilities to construct the balanced dataset MMPhysData-36k. 

\subsubsection{Multimodal Annotation}
To provide a comprehensive perception of physical content, we perform multi-granular annotation in a top-down manner, spanning semantic, geometry, and spatiotemporal trajectory.

\textit{Instance Segmentation.} Based on the subject descriptions generated by the VLM, we leverage SAM3~\cite{sam3} with promptable concept segmentation to transfer linguistic noun phrases to semantic masks. The Presence Head of SAM3 eliminates hallucinated subjects, retaining only masks for physically active subjects.

\textit{Geometry and Point Tracking.} Within the masked regions, we employ a feed-forward 3D tracker~\cite{vggt,spatialtrackerv2} to estimate scene geometry and 3D point trajectories $\{\mathbf{p}_i(s) \in \mathbb{R}^3\}_{s=1}^F$, where $s$ is the frame index. The color $\mathbf{r}_i$ of each point is assigned based on its normalized coordinates in the initial frame $(s=1)$, where we map the $(x, y, 1/z)$ coordinates into the $[0, 1]^3$ RGB space. These colors remain constant across frames to maintain temporal identity. At each frame, the 3D points are projected onto the camera plane to encode the final perceptual RGB frame.

Compared with any single perceptual modality, our unified videos provide a holistic scaffold from global semantic understanding to local geometric details, enabling physical perception for generation models.

\section{Experiments}

\subsection{Experimental Setup}

\noindent \textbf{Baselines.} To assess our joint multimodal modeling framework, we integrate MMPhysVideo within advanced T2V video diffusion models, Wan2.1-1.3B~\cite{wan2p1}, CogvideoX-5B and 2B~\cite{cogvideox}. For a comprehensive comparison, we adopt several competitive baselines with video foundation models and physical-target methods. 

\noindent \textbf{Implementation Details.} \textit{Stage $I$}. We train the teacher models on our curated MMPhysData-36k dataset for 5K steps with a global batch size of 16, using the AdamW~\cite{adamw} optimizer and a constant learning rate of $2 \times 10^{-5}$. 
\textit{Stage $II$}. The single-stream student model is initialized from the pre-trained teacher weights and distilled through representation alignment on DiT blocks with inserted control links. We finetune the student for 1K steps with a global batch size of 32 and a learning rate of $2 \times 10^{-6}$. 
For ablations, we adopt CogVideoX-2B under the \textit{Stage I} setting.

\noindent \textbf{Evaluation.} We evaluate MMPhysVideo on two physics-centric benchmarks, VideoPhy~\cite{videophy} and PhyGenBench~\cite{phygenbench}. Quantitative evaluations are conducted using VideoCon-Physics~\cite{videophy} and VideoScore2~\cite{videoscore2}, measuring \textit{Physical Consistency} (PC) and \textit{Semantic Alignment} (SA). 

\begin{table}[t!]
    \centering
    \resizebox{\linewidth}{!}{
    \begin{tabular}{l ccc cc}
    \toprule
    \multirow{2}{*}{Method} 
    & \multicolumn{3}{c|}{\textit{Physical Interaction Regime}} 
    & \multicolumn{2}{c}{\textit{Average}} \\
    \cmidrule(l{1pt}r{1pt}){2-4}
    \cmidrule(l{1pt}r{1pt}){5-6}
    & Solid-Solid & Solid-Fluid & Fluid-Fluid 
    & PC & SA \\
    \midrule
    \multicolumn{6}{l}{\textit{Video Foundation Model}} \\
    VideoCrafter2 & 32.2 & 27.4 & 29.1 & 29.7 & 50.3 \\
    DreamMachine & 21.7 & 23.3 & 18.2 & 21.8 & 57.5 \\
    LaVIE & 18.3 & 37.0 & 50.9 & 31.5 & 48.7 \\
    Cosmos-Diffusion-7B & - & - & - & 18 & 57 \\
    HunyuanVideo & 16.1 & 30.1 & 54.5 & 28.2 & 60.2 \\
    \midrule
    \multicolumn{6}{l}{\textit{Physics-aware Video Generation Model}} \\
    OmniVDiff & 4.9 & 12.3 & 23.6 & 11.0 & 45.1 \\
    PhysHPO & 20.4 & 24.7 & 42.9 & 25.9 & - \\
    PhyT2V & - & - & - & 37 & 61 \\
    WISA & - & - & - & 38 & 67 \\
    VideoREPA & 28.0 & 39.0 & 74.5 & 40.1 & 72.1 \\
    \midrule
    Wan2.1-1.3B & 16.8 & 25.3 & 41.8 & 24.4 & 68.0 \\
    \rowcolor{morandiblue}
    +MMPhysVideo & 20.3 & 30.8 & 49.1 & 29.4 & 70.6 \\
    \midrule
    CogVideoX-2B & 13.3 & 28.1 & 50.9 & 25.6 & 60.5 \\
    \rowcolor{morandiblue}
    +MMPhysVideo & 23.1 & 32.2 & 56.4 & 32.3 & 66.6 \\
    \midrule
    CogVideoX-5B & 19.6 & 33.6 & 61.8 & 32.3 & 70.0 \\
    \rowcolor{morandiblue}
    +MMPhysVideo & \textbf{32.2} & \textbf{43.2} & \textbf{78.2}
    & \textbf{44.2} & \textbf{75.0} \\
    \bottomrule
    \end{tabular}
    }
    \vspace{-2pt}
    \caption{\textbf{Results on Videophy} across three physical interaction regimes with VideoCon-Physics evaluator.}
    \label{tab:sota_videophy}
    \vspace{-8pt}
\end{table}

\begin{table}[t!]
    \centering
    \resizebox{\linewidth}{!}{
    \begin{tabular}{l cccc cc}
    \toprule
    \multirow{2}{*}{Method} 
    & \multicolumn{4}{c}{\textit{Physical Domain}}
    & \multicolumn{2}{c}{\textit{Average}} \\
    \cmidrule(lr){2-5} \cmidrule(lr){6-7}
    & Mechanics
    & Optics
    & Thermo
    & Material
    & PC
    & SA \\
    \midrule

    CogVideoX1.5-5B 
    & 27.5 & 35.0 & 40.0 & \textbf{32.5} & 33.8 & 29.4 \\

    +OmniVDiff 
    & 5.0 & 20.0 & 17.5 & 20.0 & 15.6 & 19.4 \\

    \midrule

    CogVideoX-2B 
    & 25.0 & 15.0 & 12.5 & 15.0 & 16.9 & 18.1 \\

    +PhyT2V 
    & 20.0 & 20.0 & 17.5 & 17.5 & 18.8 & 20.6 \\

    +VideoREPA 
    & 22.5 & 25.0 & 17.5 & 20.0 & 21.2 & 22.5 \\

    \rowcolor{morandiblue}
    +MMPhysVideo
    & 30.0 & 27.5 & 17.5 & 27.5
    & 25.6 & 26.3 \\

    \midrule

    CogVideoX-5B 
    & 20.0 & 35.0 & 22.5 & 17.5 & 23.8 & 21.2 \\

    +PhyT2V 
    & 20.0 & 32.5 & 22.5 & 25.0 & 25.0 & 22.5 \\

    +VideoREPA 
    & 25.0 & 32.5 & 27.5 & 25.0 & 27.5 & 25.6 \\

    \rowcolor{morandiblue}
    +MMPhysVideo
    & \textbf{32.5} & \textbf{37.5} & \textbf{40.0} & \underline{30.0}
    & \textbf{35.0} & \textbf{30.6} \\




    \bottomrule
    \end{tabular}
    }
    \vspace{-2pt}
    \caption{\textbf{Results on PhyGenBench} across four physical domains with VideoScore2 evaluator.}  
    \label{tab:sota_phygenbench}
    \vspace{-8pt}
\end{table}

\subsection{Main results}

\textbf{Quantitative Comparisons.} Results in Tab.~\ref{tab:sota_videophy} and~\ref{tab:sota_phygenbench} reveal three key observations: 
\textbf{1}) MMPhysVideo generalizes effectively across T2V backbones with different scales, consistently improving physical consistency. Specifically, it increases the Average PC scores of CogVideoX-5B by $11.9\%$ and Wan2.1-1.3B by $5.0\%$ on VideoPhy, respectively. 
\textbf{2}) MMPhysVideo achieves state-of-the-art performance on both benchmarks, surpassing specialized physics-aware methods. It outperforms PhyT2V and VideoREPA by $7.2\%$/$8.8\%$ and $10.0\%$/$7.5\%$ in Average PC on Videophy and PhyGenBench, respectively. 
\textbf{3}) MMPhysVideo significantly surpasses the multimodal baseline OmniVDiff, which adopts a stronger CogVideoX1.5-5B backbone and a training dataset over $10\times$ larger than MMPhysData-36k.

\begin{figure*}[t!]
    \centering
    \includegraphics[width=0.98\textwidth]{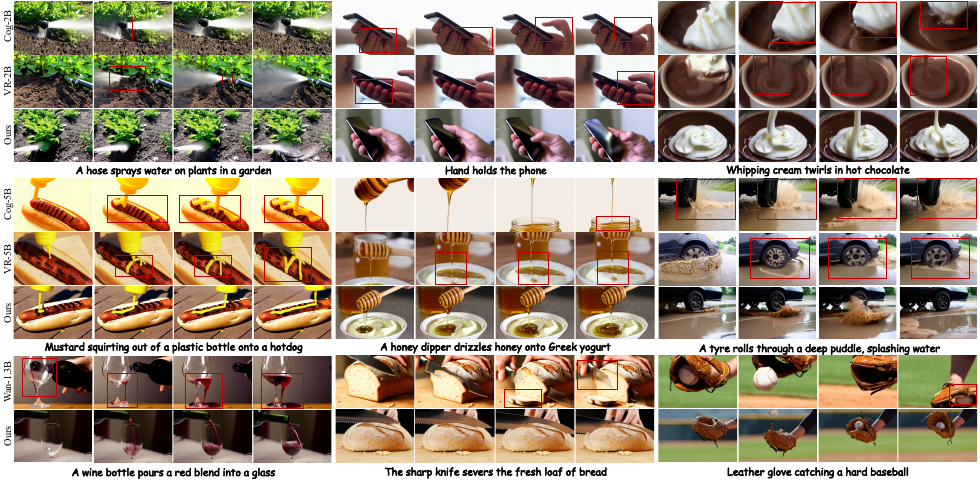}
    \vspace{-2pt}
    \caption{\textbf{Qualitative results.} We compare MMPhysVideo with backbones, CogVideoX (Cog) and Wan2.1 (Wan), alongside the advanced physics method, VideoREPA (VR). Our method exhibits clearer causal progression of physical phenomena.}
    \label{fig:case}
    \vspace{-8pt}
\end{figure*}

\noindent \textbf{Qualitative comparison.} 
Fig.~\ref{fig:case} presents qualitative comparisons among MMPhysVideo, baseline T2V generators, and the physics-aware method VideoREPA~\cite{videorepa} across diverse physical scenarios. MMPhysVideo consistently produces more physically coherent results. In the ``hand-phone'' scenario, it accurately generates anatomical structures and captures fine-grained finger-screen interactions. In the ``wine-glass'' scenario, baseline models fail to preserve object relationships and fluid dynamics, leading to object penetration and implausible liquid trajectories. In contrast, MMPhysVideo maintains structural consistency and generates physically plausible fluid motion.

\begin{figure}[t!]
    \centering
    \includegraphics[width=0.98\linewidth]{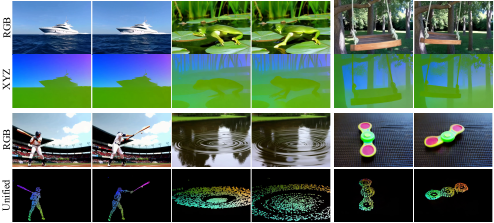}
    \vspace{-2pt}
    \caption{Visualization of RGB-perception generation, which exhibits consistent spatial structures and temporal dynamics.}
    \label{fig:percep_case}
    \vspace{-8pt}
\end{figure}

\begin{figure}[t!]
    \centering
    \includegraphics[width=0.98\linewidth]{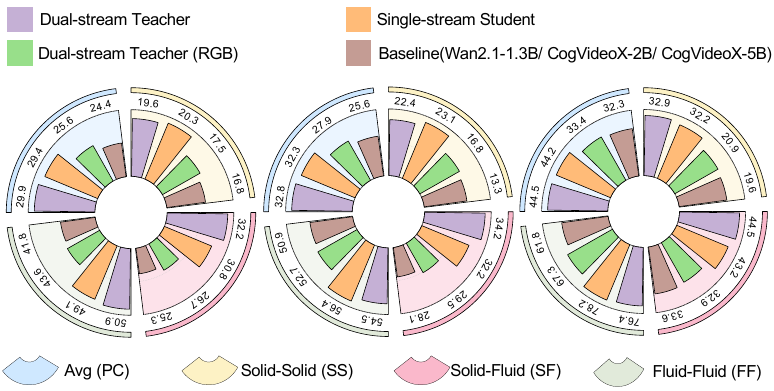}
    \vspace{-2pt}
    \caption{\textbf{Evaluation} of our representation alignment distillation based on CogVideoX-2B, CogVideoX-5B and Wan2.1-1.3B from left to right, respectively.}\label{fig:aba_distillation}
    \vspace{-8pt}
\end{figure}

\subsection{Ablations and Analysis}
\label{sec:ablation}
\noindent \textbf{Effectiveness of distillation.} In Fig.~\ref{fig:aba_distillation}, we evaluate relation-based distillation for single-stream student models across backbones. We compare the distilled student with the dual-stream teacher, a teacher variant without perceptual control signals, and the baseline. The student consistently matches the dual-stream teacher and surpasses the teacher variant, demonstrating the effectiveness of our distillation strategy. As verified in Tab.~\ref{tab:aba_arch_percep}, MSE-based distillation degrades performance by forcing the student to mimic exact feature values, whereas our method better preserves the complementary spatial and temporal relations encoded by the teacher.

\begin{figure}[t!]
    \centering
    \includegraphics[width=0.98\linewidth]{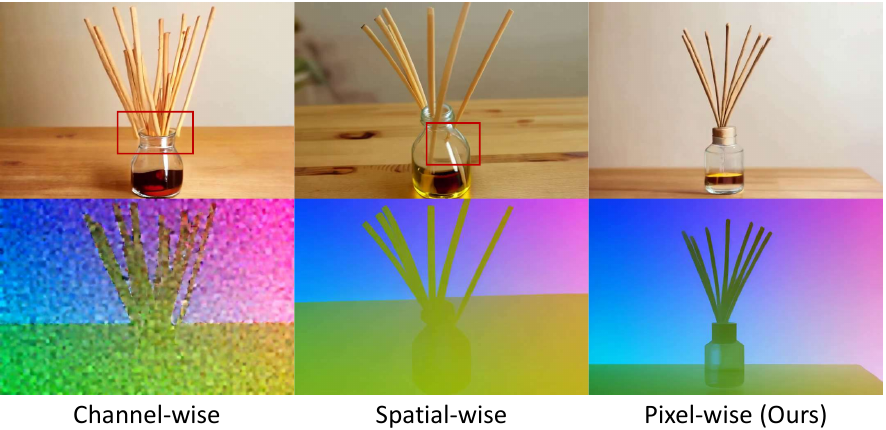}
    \vspace{-2pt}
    \caption{\textbf{Qualitative comparison of fusion strategies.} Our pixel-wise design produces more coherent results.}
    \label{fig:arch_vis_case}
    \vspace{-12pt}
\end{figure}

\noindent \textbf{Variants of architectures for joint modeling.} 
In Sec.~\ref{sec:mmphysvideo}, we introduce a dual-stream architecture with bidirectional control links between RGB and perceptual branches, enabling pixel-wise cross-modal interaction. We compare it with alternative fusion strategies, including channel-wise fusion and spatial-wise fusion, while keeping the training data and steps identical for a fair comparison. As shown in Fig.~\ref{fig:arch_vis_case} and Tab.~\ref{tab:aba_arch_percep}, our architecture achieves finer perceptual predictions and significantly improves the average PC score. Moreover, early modality fusion (e.g., channel-wise fusion) introduces severe inter-modal interference, compromising both RGB and perceptual generation quality.

\begin{table}[t!]
    \centering
    \resizebox{\linewidth}{!}{
    \begin{tabular}{l ccc cc}
    \toprule
    \multirow{2}{*}{Method} 
    & \multicolumn{3}{c|}{\textit{Physical Interaction Regime}} 
    & \multicolumn{2}{c}{\textit{Average}} \\
    \cmidrule(l{1pt}r{1pt}){2-4}
    \cmidrule(l{1pt}r{1pt}){5-6}
    & Solid-Solid & Solid-Fluid & Fluid-Fluid 
    & PC & SA \\
         \midrule
         \rowcolor{morandiblue}
         Ours & \textbf{23.1} & \textbf{32.2} & \textbf{56.4} & \textbf{32.3} & \textbf{66.6} \\
         \midrule
         \multicolumn{6}{l}{\textit{Ablations of architecture design}} \\
         Channel-wise & 18.9 & 30.1 & 36.4 & 26.4 & 62.5 \\
         Spatial-wise & 21.0 & 30.8 & 43.6 & 28.8 & 63.7 \\
         \midrule
         \multicolumn{6}{l}{\textit{Ablations of perceptual modality}} \\
         Segmentation & 19.6 & 29.5 & 38.2 & 26.7 & 65.7 \\
         Pointmap & 22.4 & 29.5 & 49.1 & 29.7 & 64.5 \\
         Point Tracking & 21.7 & 31.5 & 52.7 & 30.8 & 61.9 \\
         \midrule
         \multicolumn{6}{l}{\textit{Ablations of distillation}} \\
         Spatial Relation & 23.1 & 27.4 & 47.3  & 29.4 & 64.0 \\
         Temporal Relation & 20.3 & 28.1 & 50.9 & 28.5 & 62.8 \\
         MSE Regression & 15.4 & 19.9 & 45.5 & 22.1 & 56.4 \\
        \bottomrule
    \end{tabular}
    }
    \vspace{-2pt}
    \caption{\textbf{Ablation analysis} of joint modeling architecture design, perceptual modality and distillation variants.}\label{tab:aba_arch_percep}
    \vspace{-12pt}
\end{table}

\noindent \textbf{Analysis of perceptual modality.} To evaluate the contribution of each modality in our unified perceptual videos, we conduct a break-down ablation for the annotation in MMPhysPipe. 
Specifically, we perform separate annotations for each perceptual modality on our constructed MMPhysData-36k dataset, including segmentation, pointmap (XYZ), and point tracking. Subsequently, we only replace our unified perceptual videos with single-modality counterparts for training. Based on the experimental results presented in Tab. \ref{tab:aba_arch_percep}, we make the following observations: \textbf{1)} Our unified method outperforms all single-modality baselines, achieving the highest performance across average PC and SA scores. This indicates that our unified videos provide rich, complementary physical content for VDMs. \textbf{2)} Among the single-modality variants, instance segmentation provides the most substantial improvement in the model's semantic alignment.
Furthermore, we demonstrate the visual results of our dual-stream teacher models, which simultaneously generate RGB videos and corresponding perceptual predictions. As shown in Fig.~\ref{fig:percep_case}, the pointmap modality enables MMPhysVideo to disentangle foreground subjects from complex background in open scenes, while the integration of segmentation and point tracking enables a granular focus on physical dynamics, spanning from rigid-body motion to fluid ripples.

\section{Conclusion}

In this paper, we propose MMPhysVideo, a novel framework that enhances physical consistency in video generation via joint multimodal modeling. To achieve this, we introduce a dual-stream teacher architecture to decouple RGB and perceptual modalities, and distill it into a single-stream student through representation alignment for efficiency. We further develop MMPhysPipe, an end-to-end pipeline for curating physics-rich videos with multi-granular perceptual annotations. Extensive experiments on physics-focused benchmarks demonstrate that MMPhysVideo consistently improves advanced video generators, achieving state-of-the-art performance in physics-consistent video generation.

\bibliography{aaai2027}


\end{document}